\documentclass[conference]{IEEEtran}
\usepackage{cite}
\usepackage{graphicx}
\usepackage{pgfplots}
\usepgfplotslibrary{groupplots}
\pgfplotsset{width=\linewidth,compat=1.9}
\usepackage{url}
\usepackage{tabularx, booktabs}
\usepackage{listings}
\usepackage{xcolor}
\usepackage{makecell} 						% For centered multi-line table head
\usepackage[acronym]{glossaries}
\usepackage[nointegrals]{wasysym}			% For circle symbols (used in comparison tables)
\usepackage[shortlabels]{enumitem}			% For custom list labels (e.g., list of RQ 1 to 3)
\usepackage{csquotes}

\definecolor{light-gray}{gray}{0.95}
\lstdefinestyle{mystyle}{
    backgroundcolor=\color{light-gray},  
    basicstyle=\ttfamily\small
}

\newacronym{ABox}{ABox}{assertional box}
\newacronym{AAS}{AAS}{Asset Administration Shell}
\newacronym{LLM}{LLM}{Large Language Model}
\newacronym{ODP}{ODP}{Ontology Design Patter}
\newacronym{OWL}{OWL}{Web Ontology Language}
\newacronym{SHACL}{SHACL}{Shapes Constraint Language}
\newacronym{TBox}{TBox}{terminological box}
\newacronym{UML}{UML}{Unified Modeling Language}
\newacronym{CLI}{CLI}{Command Line Interface}
\newacronym{API}{API}{Application Programming Interface}
\newacronym{PDDL}{PDDL}{Planning Domain Definition Language}
\newacronym{SysML}{SysML}{Systems Modeling Language}

\begin{document}
\bstctlcite{IEEEexample:BSTcontrol} % To shorten reference authors

\title{Toward a Method to Generate Capability Ontologies from Natural Language Descriptions}
% Alternativen 
% Toward Automatically Generating Capability Ontologies from Natural Language Descriptions with Large Language Models
% Leveraging Large Language Models for Capability Ontology Generation
% Towards a Methodology for Extracting Capability Ontologies from Natural Language with Large Language Models
% Harnessing Large Language Models for Capability Ontology Generation from Natural Language

\author{
\IEEEauthorblockN{
    Luis Miguel Vieira da Silva\IEEEauthorrefmark{1},
    Aljosha Köcher\IEEEauthorrefmark{1},
    Felix Gehlhoff\IEEEauthorrefmark{1}, 
    Alexander Fay\IEEEauthorrefmark{2}
}
\IEEEauthorblockA{
\IEEEauthorrefmark{1}Institute of Automation\\
Helmut Schmidt University, Hamburg, Germany\\
Email: \{miguel.vieira, aljosha.koecher, felix.gehlhoff\}@hsu-hh.de\\}

\IEEEauthorblockA{
		\IEEEauthorrefmark{2} Chair of Automation \\
		Ruhr University, Bochum, Germany \\
		Email: alexander.fay@rub.de}
}

\maketitle

\begin{abstract}
To achieve a flexible and adaptable system, capability ontologies are increasingly leveraged to describe functions in a machine-interpretable way. 
However, modeling such complex ontological descriptions is still a manual and error-prone task that requires a significant amount of effort and ontology expertise. 
This contribution presents an innovative method to automate capability ontology modeling using \glspl{LLM}, which have proven to be well suited for such tasks. 
Our approach requires only a natural language description of a capability, which is then automatically inserted into a predefined prompt using a few-shot prompting technique.  
After prompting an \gls{LLM}, the resulting capability ontology is automatically verified through various steps in a loop with the \gls{LLM} to check the overall correctness of the capability ontology. 
First, a syntax check is performed, then a check for contradictions, and finally a check for hallucinations and missing ontology elements. 
Our method greatly reduces manual effort, as only the initial natural language description and a final human review and possible correction are necessary, thereby streamlining the capability ontology generation process. 

\glsreset{LLM}
\end{abstract}


% Abstract + Intro max. 1 Seite 
\section{Introduction}
\label{sec:introduction}
In today's rapidly evolving manufacturing landscape, companies must adapt quickly to shifting demands, necessitating highly flexible systems.  
Consequently, it is imperative to select and modify systems and functions as needed. 
To facilitate such adaptability it is crucial to provide machine-interpretable descriptions of these functions.   
In this regard, the concepts of \emph{capabilities} and \emph{skills} have emerged as important concepts in recent years, due to the standardization efforts by working groups \emph{Plattform Industrie 4.0}\footnote{\url{https://www.plattform-i40.de/IP/Redaktion/EN/Downloads/Publikation/CapabilitiesSkillsServices.html}} and \emph{IDTA}\footnote{\url{https://github.com/admin-shell-io/submodel-templates/tree/main/development/Capability/1/0}}.
Capabilities are defined as an implementation-independent specification of a function, while skills are executable implementations of a function specified by a capability \cite{KBH+_AReferenceModelfor_15.09.2022b}. 
However, the manual specification of capabilities and skills is a tedious effort. 
Therefore, the automated generation of such specifications is a topic of current research.

To generate machine-interpretable models of capabilities and skills, ontologies are predominantly used due to their benefits in knowledge representation (e.g., querying, reasoning, constraint-checking).  
As formal information models, ontologies define a set of concepts, delineate their meanings, and specify the relations between them \cite{Usc_Knowledgelevelmodelling:concepts_1998}.
%Ontologies can be divided into a \gls{TBox} and an \gls{ABox}. While a \gls{TBox} entails class knowledge about a problem domain, an \gls{ABox} contains factual knowledge of one specific problem \cite{BCM+_TheDescriptionLogicHandbook_2007}.
However, the generation of model instances for existing capability ontologies demands considerable ontology expertise and significant effort. 
Although there are approaches that automatically generate parts of these ontologies from existing information (e.g., \cite{Kocher.08.09.202011.09.2020}), the modeling of capabilities remains a manual, labor-intensive, and error-prone task. 

\glspl{LLM} are promising for the automated generation of capability ontologies, as they show great potential for natural language processing tasks such as translation and code generation \cite{CWW+_ASurveyonEvaluation_2024}. 
%\glspl{LLM} are advanced computational models that primarily utilize transformer architecture to generate text \cite{CWW+_ASurveyonEvaluation_2024}. 
In our previous study \cite{VKG+_LLMsGeneratingCapability_2024} we compared different \glspl{LLM} and prompting techniques for generating a capability ontology and confirmed their suitability for this task.  
Prompting techniques stem from the field of prompt engineering, which involves crafting prompts to interact with \glspl{LLM} in a way that maximizes the accuracy and relevance of their responses. 
One notable prompting technique is \emph{few-shot prompting}, which leverages \emph{in-context learning} by providing context in the form of a number of examples within a prompt \cite{BMR+_LanguageModelsareFewShot_28.05.2020}.
% to guide an \gls{LLM} towards generating more coherent responses \cite{BMR+_LanguageModelsareFewShot_28.05.2020}.

This article proposes a method utilizing \glspl{LLM} to automate the generation of capability ontologies, thereby minimizing the manual effort.
However, \glspl{LLM} may cause \emph{hallucinations} by generating information that is factually incorrect, invented, or irrelevant to the given natural language input \cite{CWW+_ASurveyonEvaluation_2024}. 
Therefore, our method incorporates an automated, systematic verification and refinement of the resulting capability ontology in order to enable its integration into applications such as manufacturing execution systems. 

%Therefore, the following research sub goals are targeted to be automated: 
%\begin{enumerate}[start=1,leftmargin=*,label=\textbf{RG \arabic*:}]
%    \item Generation of capability ontologies with \glspl{LLM} and well-proofed prompting techniques  
%    \item Verification of resulting capability ontologies
%\end{enumerate}

The remainder of this paper is structured as follows: Section~\ref{sec:relatedWork} reviews existing contributions to capability engineering methods and studies covering the use of \glspl{LLM} to generate machine-interpretable models. 
Section~\ref{sec:method} details our proposed method and Section~\ref{sec:implementaiton} gives an overview of the implementation. The paper concludes with a discussion of our findings and future research directions in Section~\ref{sec:conclusion}.

% 0,5 Seite
\section{Related Work} 
\label{sec:relatedWork}
% Eher raus, zu allgemein:
% In \cite{Jarvenpaa.2019} a systematic development process for an ontology to describe capabilities of manufacturing resources is proposed. 
% This approach is based on the ontology engineering methodology of \cite{Sure.2009} and consists of five phases: feasibility study, kickoff, refinement, evaluation as well as application and evolution. 
% The focus of this method is on modeling the \gls{TBox}. 
% An \gls{ABox} is modeled in the evaluation phase, but no methodological approach or automated process is presented to support this activity. 

In \cite{Kocher.08.09.202011.09.2020}, we present a method to create the various aspects of a capability and skill ontology from existing engineering artifacts in order to reduce the high effort required for manual ontology creation. 
Using a provided framework, the skill aspect is created automatically using source code of the skill behavior.
For the capability aspect, a semi-automated approach using graphical modeling is used as there are no engineering artifacts describing functions in a structured manner. 
But even though that approach creates guidance in creating a capability model, it still requires some manual efforts for creating a graphical model. Furthermore, some parts of the capability aspect are not covered (e.g., constraints).
% Therefore, the graphical modeling tool fpb.js in \cite{Nabizada.2020} is used to manually model capabilities as a process with inputs and outputs based on the formalized process description of the VDI 3682 standard \cite{VDI3682}.
% and reduce the effort for ontology creation. 
% A capability can be modeled as a process with inputs and outputs. 
% The resulting JSON file can be automatically mapped to the ontology. 
% This modeling does not allow the description of properties and constraints of capabilities and their inputs or outputs. 
% Nor can constraints between inputs and outputs or between capabilities be modeled.

% Raus, zu allgemein
% The \emph{Chowlk} framework presented in \cite{Chavez.2022} offers the possibility to visually model a \gls{TBox} based on \gls{UML} instead of using cumbersome ontology editors.
% With Chowlk, a created diagram is automatically transformed into an \gls{OWL} ontology. 
% This framework is primarily intended to reduce the effort required to develop a \gls{TBox} of an ontology for a specific domain, but also allows the creation of an \gls{ABox}. 
% The manual effort required to create an \gls{ABox} remains high and a user still needs to be an ontology expert to be able to use the framework and understand which elements to select. 

Besides ontologies, the \gls{AAS} is also used to create capability models using an existing submodel template.
In the context of the \gls{AAS}, there are also initial approaches to support modeling processes: 
One such approach is shown in \cite{Huang.2021}, which presents a modeling process based on a graphical modeling framework. Similar to \cite{Kocher.08.09.202011.09.2020}, this approach also eases the creation of a capability model by giving users an easy-to-use modeling environment. Models created in this environment are automatically translated to the actual \gls{AAS} model.
Another work aiming to automate the creation of \gls{AAS} models is presented in \cite{Xia.25.03.2024}. In \cite{Xia.25.03.2024}, \glspl{LLM} are used to generate instances of \gls{AAS} from textual technical data. For this purpose, a so-called \emph{semantic node} is introduced to capture the semantic essence of textual data. Multiple \glspl{LLM} and system designs were evaluated and the results are promising, showing an effective generation rate of 62--79\%. Manual effort is only required to verify the results, for which no method is provided. 

%In \cite{Trajanoska.08.05.2023}, it is more generally investigated to what extent an \gls{LLM} like ChatGPT and REBEL, a model specifically trained for relation extraction, can be used to extract relations from unstructured text. Both models generate roughly the same number of triples. Comparatively simple prompts were used for ChatGPT and the authors point out that better results can be expected with prompt engineering.
%Unfortunately, neither the prompts used nor the results of \cite{Trajanoska.08.05.2023} are publicly available. In addition, the evaluation was only carried out manually using informal criteria (e.g., "Triples should be concise"). 
%In contrast, we aim to use automated evaluation techniques based on reasoning and constraint checking as well as guided improvements through repeated prompting, so-called \emph{backprompting}.

The authors of \cite{Funk.18.09.2023} present an approach to automatically create a concept hierarchy using GPT-3.5 Turbo. Starting from a given concept, e.g., \emph{animal}, ChatGPT is repeatedly asked for subconcepts until a complete taxonomy is created.
The results are considered promising, despite some cases of hallucination and incompleteness. However, as the created ontology is only a taxonomy, complex relations like dependencies or constraints found in our capability ontology are not generated. Furthermore, verification is achieved through manual prompts. With our method, we aim to automate back-prompting.

%In \cite{Meyer.13.07.2023}, a variety of experiments is conducted with ChatGPT to investigate the extent to which \glspl{LLM} can be used for knowledge graph engineering.
%These experiments include direct questions about a knowledge graph, generation of SPARQL queries as well as knowledge graph construction.
%Although the authors report promising results, they also state that some of the results contain errors - without going into these errors in detail.
%Rather simple prompts are used throughout \cite{Meyer.13.07.2023} and no different prompting techniques are compared to obtain a better solution. Furthermore, there is no method to integrate the \gls{LLM} into a workflow and verify the results. 

In \cite{BabaeiGiglou.2023} the \emph{LLM4OL} approach, which uses \glspl{LLM} for ontology learning, is presented. Three types of tasks are examined: type discovery, recognizing taxonomies and discovery of non-taxonomic relations. A zero-shot prompting technique and 11 different \glspl{LLM} are used to study these tasks for different knowledge domains. 
The authors of \cite{BabaeiGiglou.2023} conclude that \glspl{LLM} are suitable assistants in ontology learning, but that task-specific finetuning is required to achieve practically viable solutions. However, only zero-shot prompts are used, so it is not clear to what extent prompt engineering alone would have led to better results.

% --------- {Guan.2023} nimmt backprompting vor -------------------------------
\glspl{LLM} are not only used to create ontologies, but also in other areas where manually creating models is known to be time-consuming and complicated. 
Ref.~\cite{Guan.2023} is an approach to generate AI planning problems in the language \emph{\gls{PDDL}} from natural language using \glspl{LLM}. 
Both the description of the domain and the problem to solve are generated. The contribution describes that \glspl{LLM} cannot perform planning or reasoning themselves, but can support planning by generating the planning problem in existing languages. Conventional planners are then used to solve the planning problem. 
In the case of an incorrect planning problem, \gls{PDDL} validators are used to check the syntax and human experts are used to check for factual errors and incompleteness. The findings are input as an additional prompt to the \gls{LLM}, which then makes corrections. Both GPT-4 and GPT-3.5 Turbo are used and few-shot is used as a prompting technique. Even though the approach in \cite{Guan.2023} is built for \gls{PDDL}, a different modeling language from a different domain, our idea is close to \cite{Guan.2023} because it describes a method that uses the formalism of the generated model to perform verification and correction.

A similar method is presented in \cite{ApSu_SystemArchitectsArenot_22120242232024} for the domain of systems engineering. In this paper, Apvrille and Sultan show how GPT can be used to automatically generate valid SysML diagrams from natural-language specifications. After inputting knowledge on systems engineering and diagram types, the specification is separated into multiple prompts and a JSON-formatted answer is requested. Answers are automatically parsed into SysML diagrams and analyzed for correctness. On errors, additional prompts are issued automatically or based on user interaction. In an evaluation against a conventional, manual modeling workflow, the developed framework is shown to achieve better results in a significantly shorter time. However, the authors of \cite{ApSu_SystemArchitectsArenot_22120242232024} argue that their framework is not a direct replacement for systems engineers, but instead needs to be combined with human expertise, especially for complex real-world systems.
Just like \cite{Guan.2023}, the approach of \cite{ApSu_SystemArchitectsArenot_22120242232024} has some commonalities with our approach, but deals with a different modeling domain and language.

%We have shown that there is no suitable methodology to simplify the modeling of capabilities and that \glspl{LLM} are suitable for the generation of models, but in current contributions generating ontologies there is a lack of a proper method for verifying the partially incorrect results of a \gls{LLM} as well as a study for determining an appropriate \gls{LLM} and prompting technique. 

% 2 Seiten
\section{LLM4Cap Method}
\label{sec:method}
% 1. Aufstellen der NL-TaskDescription. Inklusive aller Properties, Constraints usw.
% 2. Prompt wird automatisch generiert durch Befüllen eines Templates
% 3. Request an LLM, Ontologie liegt vor. dann Checks
% 4. Syntax-Validation. Wenn fail, dann Back-Prompt. Wenn dann immer noch fehlschlägt: Experte
% 5. Reasoning. Wenn fehlschlägt: Experte soll es sich anschauen mit den Explanations (eventuell Backprompting)
% 6. SHACL: Wenn Fehler: Backprompting
% (7. SPARQL, Competency Questions. Könnte aber auch in SHACL machen. Müssten durch user angelegt werden)
% 8. Check du Experten hinsichtlich Vollständigkeit
% Gesamt-Ergebnis: Enorme Arbeitserleichterung für Experten, aber kein einfach zu nutzender Automatismus
% 
\begin{figure*}[htb]
    \centering
    \includegraphics[width=\linewidth]{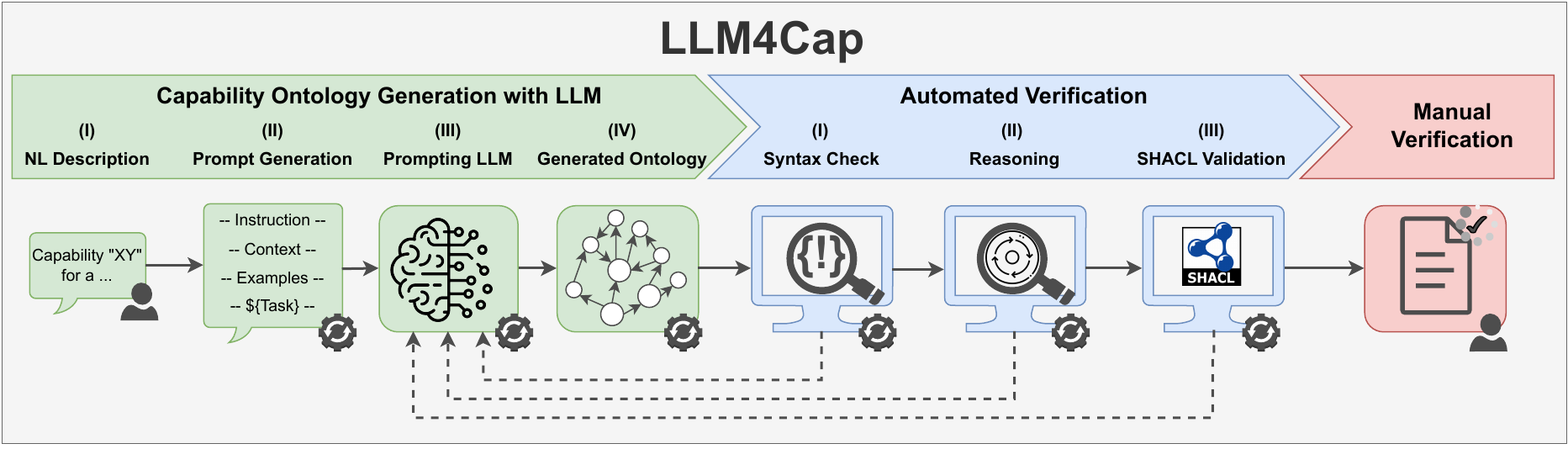}
    \caption{Overview of the proposed approach LLM4Cap: A natural language description of a capability is transformed into a capability ontology through an \gls{LLM} and then verified in four steps.}
    \label{fig:concept}
\end{figure*}

The creation of model instances for specific capabilities based on an existing capability ontology has been a time-consuming manual process that requires a great amount of expertise. 
To address this, we propose the \emph{LLM4Cap} method (see Figure~\ref{fig:concept}), which on the one hand automatically generates these model instances with the help of \glspl{LLM} and on the other hand automatically checks the generated results.  
A detailed step-by-step explanation of the method presented in Figure~\ref{fig:concept} follows in the next subsection.

\subsection{Capability Ontology Generation with an \gls{LLM}}
First, a description of the capability to be generated must be provided in natural language by a human (see \textbf{NL Description} in Figure~\ref{fig:concept}). 
This description is the only manual step required during the generation process. 
The natural language description should clarify the effect of the capability within the physical or virtual world, detailing the inputs and outputs of a capability, such as products or information, as well as their properties (e.g., position). 
Each property must be clearly categorized based on its purpose --- whether it represents a current value, a requirement for applying the capability, an assured property after applying the capability, or a variable parameter. 
Furthermore, existing dependencies or constraints among these properties should be thoroughly outlined.
It is important to comprehensively describe all necessary elements or details of the capability to ensure its accurate modeling. Otherwise, missing elements cannot be modeled.    

The description of the capability to be generated is used as input in the next step of \textbf{Prompt Generation} as shown in Figure~\ref{fig:concept}. 
The prompt is predefined and consists of the following aspects: 
\begin{itemize}
    \item Short and concise \emph{instruction} to translate the following task description describing a capability into an \gls{OWL} ontology in Turtle syntax based on the capability ontology from the context description.  
    \item \emph{Context} is established by the capability ontology \emph{CaSk}\footnote{https://github.com/caskade-automation/cask}, which is based on industry standards and can be used to describe systems, their capabilities, and executable skills. The CaSk ontology is detailed in \cite{KHV+_AFormalCapabilityand_9820209112020}.   
    This context furnishes the \gls{LLM} with possible classes and relations for the purpose of capability modeling.  
    \item \emph{Examples} according to the few-shot prompting technique. 
    In our previous study in \cite{VKG+_LLMsGeneratingCapability_2024}, we examined the suitability of \glspl{LLM} for generating capabilities and compared different prompting techniques --- zero-shot, one-shot and few-shot prompting technique --- with few-shot providing the best results. 
    Each example consists of the capability description in natural language as input and the resulting capability ontology as an expected output. 
    The three examples used are a coffee-making capability, a simple mathematical operation with multiplication and a more complex capability of distillation.  
    \item \emph{Task} description serves as a placeholder for a specific capability to create and is automatically replaced by the natural language text input from the previous step.  
\end{itemize}  

In the next step \textbf{Prompting LLM} the generated prompt is sent to the \gls{LLM}. 
The \gls{LLM} is then tasked with generating the capability ontology in the last step (\textbf{Generate Ontology}).  
To ensure deterministic and reliable solutions, the parameter \emph{temperature} is set to zero.
Although our study \cite{VKG+_LLMsGeneratingCapability_2024} shows that \glspl{LLM} deliver good results, it also shows that \glspl{LLM} hallucinate and may not always provide completely correct results. 
Therefore, a verification of the generated ontology is imperative for its subsequent integration into applications.

\subsection{Verification of \gls{LLM}-Generated Ontology}
The verification process comprises three automated steps. 
Should an error be detected in any of these steps, the error is automatically backprompted to \gls{LLM} for correction, as \glspl{LLM} show good results in correcting errors with the inclusion of feedback. 
If an error recurs in the same step, the result is passed directly to the human for manual verification, thus preventing a potentially infinite cycle of error generation within the capability ontology generation process. 

The initial step of the verification process is the \textbf{Syntax Check} to ensure the structural and syntactical correctness of the ontology. 
The syntax check identifies any character errors, missing prefixes, or violations of formal syntax rules defined by the \gls{OWL} standard. 

Following the provision of a syntactically valid ontology, the logical consistency of the ontology is evaluated in the next step \textbf{Reasoning} by employing a standard, off-the-shelf reasoner. 
Reasoning is used to infer new knowledge from the modeled instances, thereby identifying contradictions. 
These contradictions indicate that the \gls{LLM} does not adhere to the predefined rules governing classes and relations within the capability ontology, resulting in an inconsistent ontology. 
For example, instances that belong to two disjoint classes are identified as contradictory. 
This step is critical for verifying the accurate modeling of instances within the defined ontology. 
%If there is a consistent ontology, the ontology is described correctly in general, but is possibly not correct for a capability ontology. 

For a more specific check of the ontology with regard to capabilities, further and more specific rules are used in the last step \textbf{\gls{SHACL} Validation} (see Figure~\ref{fig:concept}).  
\gls{SHACL} is used to describe constraints that ensure the presence or absence of certain information within an ontology. 
With a so-called \emph{closed shape} only relations explicitly defined within the \gls{SHACL} constraint are permitted to exist. 
Otherwise, invalid relations are identified and the ontology violates the constraints --- indicating hallucination of an \gls{LLM}.  
Moreover, \gls{SHACL} constraints provide a preliminary check for incompleteness by signaling the absence of mandatory relations for a comprehensive modeled capability.  

After a successful automated verification, the final necessary step is \textbf{Manual Verification}, conducted by a domain expert. 
This is necessary to ensure that the generated ontology aligns with the required capability elements.
It is not possible to verify the full extent of a specific capability automatically, and thus this manual step remains indispensable.

\section{Implementation}
\label{sec:implementaiton}
LLM4Cap is implemented as a Java framework. The decision in favor of Java was made in particular due to the availability of reasoners and SHACL validators. Especially in the area of reasoners, there is only a very limited selection for Python --- which is typically used in the context of \glspl{LLM}.

%LLM4Cap is implemented as a Java framework and offers both a \gls{CLI} and a RESTful \gls{API} for user interaction. 
LLM4Cap supports capability ontology generation using either \emph{GPT-4}\footnote{https://openai.com/index/gpt-4/} or \emph{Claude 3}\footnote{https://claude.ai}, depending on user selection. Claude is selected by default as it performed best in our previous study \cite{VKG+_LLMsGeneratingCapability_2024}. 
For prompting both \glspl{LLM} their available RESTful \glspl{API} are used.
The generated ontology is verified using the \emph{Apache Jena} framework\footnote{https://jena.apache.org/}, which is selected for its great possibilities in handling ontologies.  
Apache Jena supports various reasoners, including \emph{Pellet}\footnote{https://github.com/stardog-union/pellet}, and enables the integration of SHACL constraints. 

LLM4Cap is provided as a library to be integrated into other Java projects, as a standalone \gls{CLI} to be used directly by users as well as a RESTful \gls{API}. This RESTful \gls{API} can be used to integrate LLM4Cap into other, web-based systems, where interactions via REST \glspl{API} is common. Users can input the natural language description of the desired capability either through a text file or directly as text input. 
The components of our framework are accessible online at https://github.com/CaSkade-Automation/LLM4Cap.   

% 0,5 Seite
\section{Conclusion}
\label{sec:conclusion}
In this contribution, we presented a method to automatically generate capability ontologies from natural language descriptions with \glspl{LLM} and to automatically verify and correct the results.
The use of \glspl{LLM} greatly reduces the effort and expertise needed to create such a capability ontology. 

Future work should evaluate the prompt by testing more prompting-techniques, optimizing the natural language capability description and improving the CaSk ontology to make it better understandable for the \gls{LLM}. 
The natural language description could be improved by defining a more structured format instead of free text. And the ontology could be improved by adding labels, comments and similar annotation properties. These elements were originally intended to foster human understanding, but are now a way of providing \glspl{LLM} with more context information.

As we have shown in more detail in \cite{VKG+_LLMsGeneratingCapability_2024}, due to the size of the ontologies used, our prompts have a very high token count. This leads to comparatively high costs per prompt and also limits the selection of possible \glspl{LLM} to the two with the biggest context window.
More efficient ways to integrate context information by using embedding techniques like the one presented in \cite{CHJ+_OWL2Vec*:embeddingofOWL_2021} are thus worth investigating.

\section*{Acknowledgment}
This research in the RIVA project is funded by dtec.bw – Digitalization and Technology Research Center of the Bundeswehr. dtec.bw is funded by the European Union – NextGenerationEU

\bibliographystyle{./bibliography/IEEEtran}
\bibliography{./bibliography/references} 

\begin{thebibliography}{10}
\providecommand{\url}[1]{#1}
\csname url@samestyle\endcsname
\providecommand{\newblock}{\relax}
\providecommand{\bibinfo}[2]{#2}
\providecommand{\BIBentrySTDinterwordspacing}{\spaceskip=0pt\relax}
\providecommand{\BIBentryALTinterwordstretchfactor}{4}
\providecommand{\BIBentryALTinterwordspacing}{\spaceskip=\fontdimen2\font plus
\BIBentryALTinterwordstretchfactor\fontdimen3\font minus
  \fontdimen4\font\relax}
\providecommand{\BIBforeignlanguage}[2]{{%
\expandafter\ifx\csname l@#1\endcsname\relax
\typeout{** WARNING: IEEEtran.bst: No hyphenation pattern has been}%
\typeout{** loaded for the language `#1'. Using the pattern for}%
\typeout{** the default language instead.}%
\else
\language=\csname l@#1\endcsname
\fi
#2}}
\providecommand{\BIBdecl}{\relax}
\BIBdecl

\bibitem{KBH+_AReferenceModelfor_15.09.2022b}
A.~K{\"o}cher, A.~Belyaev \emph{et~al.}, ``{A Reference Model for Common
  Understanding of Capabilities and Skills in Manufacturing},'' \emph{{at -
  Automatisierungstechnik}}, no.~2, 2023.

\bibitem{Usc_Knowledgelevelmodelling:concepts_1998}
M.~Uschold, ``Knowledge level modelling: concepts and terminology,'' \emph{The
  Knowledge Engineering Review}, vol.~13, no.~1, pp. 5--29, 1998.

\bibitem{Kocher.08.09.202011.09.2020}
A.~Köcher, C.~Hildebrandt \emph{et~al.}, ``{Automating the Development of
  Machine Skills and their Semantic Description},'' in \emph{2020 25th IEEE
  International Conference on Emerging Technologies and Factory Automation
  (ETFA)}.\hskip 1em plus 0.5em minus 0.4em\relax IEEE, 08.09.2020 -
  11.09.2020, pp. 1013--1018.

\bibitem{CWW+_ASurveyonEvaluation_2024}
Y.~Chang, X.~Wang \emph{et~al.}, ``{A Survey on Evaluation of Large Language
  Models},'' \emph{{ACM Transactions on Intelligent Systems and Technology}},
  vol.~15, no.~3, pp. 1--45, 2024.

\bibitem{VKG+_LLMsGeneratingCapability_2024}
L.~M. Vieira~da Silva, A.~Köcher \emph{et~al.}, ``{On the Use of Large
  Language Models to Generate Capability Ontologies},'' in \emph{accepted for
  publication at 2024 29th IEEE ETFA}, 2024.

\bibitem{BMR+_LanguageModelsareFewShot_28.05.2020}
T.~Brown, B.~Mann \emph{et~al.}, ``{Language Models are Few-Shot Learners},''
  in \emph{Advances in Neural Information Processing Systems}, H.~Larochelle,
  M.~Ranzato \emph{et~al.}, Eds., vol.~33.\hskip 1em plus 0.5em minus
  0.4em\relax Curran Associates, Inc., 2020, pp. 1877--1901.

\bibitem{Huang.2021}
Y.~Huang, S.~Dhouib, and J.~Malenfant, ``{An AAS Modeling Tool for
  Capability-Based Engineering of Flexible Production Lines},'' in \emph{IECON
  2021 - 47th Annual Conference of the IEEE Industrial Electronics
  Society}.\hskip 1em plus 0.5em minus 0.4em\relax Piscataway, NJ: IEEE, 2021,
  pp. 1--6.

\bibitem{Xia.25.03.2024}
Y.~Xia, Z.~Xiao \emph{et~al.}, ``{Generation of Asset Administration Shell With
  Large Language Model Agents: Toward Semantic Interoperability in Digital
  Twins in the Context of Industry 4.0},'' \emph{IEEE Access}, vol.~12, pp.
  84\,863--84\,877, 2024.

\bibitem{Funk.18.09.2023}
\BIBentryALTinterwordspacing
M.~Funk, S.~Hosemann \emph{et~al.}, ``{Towards Ontology Construction with
  Language Models}.'' [Online]. Available:
  \url{http://arxiv.org/pdf/2309.09898}
\BIBentrySTDinterwordspacing

\bibitem{BabaeiGiglou.2023}
H.~{Babaei Giglou}, J.~D'Souza, and S.~Auer, ``{LLMs4OL: Large Language Models
  for Ontology Learning},'' in \emph{The semantic web - ISWC 2023}, ser.
  Lecture Notes in Computer Science, T.~R. Payne, V.~Presutti \emph{et~al.},
  Eds.\hskip 1em plus 0.5em minus 0.4em\relax Cham: Springer, 2023, vol. 14265,
  pp. 408--427.

\bibitem{Guan.2023}
L.~Guan, K.~Valmeekam \emph{et~al.}, ``{Leveraging Pre-trained Large Language
  Models to Construct and Utilize World Models for Model-based Task
  Planning},'' \emph{Advances in Neural Information Processing Systems},
  vol.~36, pp. 79\,081--79\,094, 2023.

\bibitem{ApSu_SystemArchitectsArenot_22120242232024}
L.~Apvrille and B.~Sultan, ``{System Architects Are not Alone Anymore:
  Automatic System Modeling with AI},'' in \emph{{Proceedings of the 12th
  International Conference on Model-Based Software and Systems
  Engineering}}.\hskip 1em plus 0.5em minus 0.4em\relax {SCITEPRESS - Science
  and Technology Publications}, 2024, pp. 27--38.

\bibitem{KHV+_AFormalCapabilityand_9820209112020}
A.~K{\"o}cher, C.~Hildebrandt \emph{et~al.}, ``{A Formal Capability and Skill
  Model for Use in Plug and Produce Scenarios},'' in \emph{{2020 25th IEEE
  International Conference on Emerging Technologies and Factory Automation
  (ETFA)}}.\hskip 1em plus 0.5em minus 0.4em\relax IEEE, 9/8/2020 - 9/11/2020,
  pp. 1663--1670.

\bibitem{CHJ+_OWL2Vec*:embeddingofOWL_2021}
J.~Chen, P.~Hu \emph{et~al.}, ``{OWL2Vec*: embedding of OWL ontologies},''
  \emph{{Machine Learning}}, vol. 110, no.~7, pp. 1813--1845, 2021, pII: 5997.

\end{thebibliography}

\end{document}